# Multiphysics discovery with moving boundaries using Ensemble SINDy and Peridynamic Differential Operator

Ali Can Bekar[a], Ehsan Haghighat[b], Erdogan Madenci[a,1]
[a]*The University of Arizona, Tucson, AZ 85721, USA*
[b]*Massachusetts Institute of Technology, Cambridge, MA 02139, USA*

**Abstract:** This study proposes a novel framework for learning the underlying physics of phenomena with moving boundaries. The proposed approach combines Ensemble SINDy and Peridynamic Differential Operator (PDDO) and imposes an inductive bias assuming the moving boundary physics evolve in its own corotational coordinate system. The robustness of the approach is demonstrated by considering various levels of noise in the measured data using the 2D Fisher-Stefan model. The confidence intervals of recovered coefficients are listed, and the uncertainties of the moving boundary positions are depicted by obtaining the solutions with the recovered coefficients. Although the main focus of this study is the Fisher-Stefan model, the proposed approach is applicable to any type of moving boundary problem with a smooth moving boundary front without a mushy region. The code and data for this framework is available at: https://github.com/alicanbekar/MB_PDDO-SINDy.

## 1. Introduction

Moving boundary problems are ubiquitous in engineering and biological systems such as melting or solidification [1,2], tumor growth and wound healing [3], free surface flows [4], and electrophotography [5]. Usually, moving boundaries split the solution domain with different governing PDEs for each subdomain. Also, interface of the split regions obeys a different governing equation [5–7], often expressed in its own coordinate system. Therefore, the moving boundary of the domain is solved in addition to the field variables. These features make physical systems with moving boundaries an intriguing and challenging case for learning their governing PDEs from field measurements.

Stefan type problems represent a suitable model for learning in the context of moving boundaries. Stefan problem describes heat transfer and interface evolution between a liquid and a solid domain. They have long been used as a benchmark problem for numerical solvers with a rich literature [8]. Various implicit and explicit solvers were developed to solve forward and inverse Stefan problem [1,9,10]. Recently, Wang and Perdikaris [11] proposed a Physics-Informed Neural Network (PINN) solver for forward and inverse solutions of a Stefan type problem. They assume that either physics or data about the system are partially available, such as certain terms in the PDEs or sparse measurements. Hence, their approach is not applicable if the terms in the governing equation are not known. When the governing equation of the system

---

[1] Corresponding author. Tel.: +1 520 621 6113.
*E-mail addresses:* acbekar@arizona.edu (A. C. Bekar), ehsanh@mit.edu (E. Haghighat), madenci@email.arizona.edu (E. Madenci)

is unknown and the only available information is the measured data, reliable and physically consistent predictions can be made by discovering the underlying physics of the system from the field measurements. Purely data driven methods such as Dynamic Mode Decomposition (DMD) also offers successful short term predictions [12]. However, model discovery reveals interpretable models with more generalization capabilities [13].

Model discovery particularly in the presence of noise is a challenging task. Bongard and Lipson used a symbolic regression approach to discover governing equations of dynamical systems from measurements [14]. Schaeffer introduced the concept of sparse learning of PDEs [15]. The key idea is to numerically calculate the time derivatives of the field data and create a matrix that comprises the spatial derivatives of the field data and subsequently cast the problem as a sparse regression between the two steps. Sparse regression is the mechanism to inflict parsimony. Schaeffer solved sparse optimization with the Douglas-Rachford algorithm [15]. This makes the framework robust to noise in time derivatives of the field variable. Brunton et al. [16] introduced Sparse Identification of Nonlinear Dynamics (SINDy). SINDy is a versatile framework with an efficient sparse optimization algorithm. They convert the nonlinear model identification to linear system of equations similar to [15] and solve for the sparse optimization by sequentially thresholding the least squares solutions to promote sparsity. In their influential work, Zhang and Schaeffer [17] analyzed the convergence behavior of SINDy, strengthening the theoretical foundation of the method. Subsequently, SINDy has been widely adopted in the field of learning PDEs and applied to a wide range of problems [18–21]. Inspired by [22], Messenger and Bortz extended the work to leverage the weak formulation of ODEs and PDEs (Weak SINDy) [23,24]. This method eliminates the pointwise approximation of derivatives using the weak form integral and significantly increases noise tolerance of SINDy algorithm. Fasel et al. [25] combined ensemble learning with the SINDy algorithm and introduced Ensemble SINDy. They demonstrated the capabilities of Ensemble SINDy by recovering the coefficients of several PDEs from noisy and scarce measurements. While Ensemble SINDy may not be as effective as Weak SINDy in handling noise, it can be used for experiments with moving boundaries because of the pointwise and corotational nature of the the problem. Furthermore, Ensemble SINDy can be combined with Weak SINDY to enhance its performance.

Combining Ensemble SINDy and Peridynamic Differential Operator (PDDO) introduced by Madenci et al. [26], we propose a discovery framework for learning dynamics of moving boundaries. We assume the moving boundary physics is governed in its own corotational coordinate system, with normal and tangential directions to the boundary, and its derivatives are calculated using the PDDO. The PDDO enables differentiation through integration and does not require uniform sensor placement. It simply considers the interaction between neighboring points for the evaluation of derivatives, it can be used to calculate derivatives in any coordinate system by a straightforward modification and it integrates well with the existing PDE discovery methods [27]. There appears no study in the open literature addressing the discovery of governing equations of moving boundaries.

This study is organized as follows. In Section 2, we explain the Ensemble SINDy. In Section 3, we briefly describe the multiphysics PDE model, Fisher-Stefan model and the numerical experiment to create training data synthetically. In Section 4, we present the results and performance of the proposed learning framework for different levels of noise in experimental data. Finally, we discuss the results and summarize the main conclusions in Section 5.

## 2. Ensemble SINDy

Introduced by Brunton et al. [16], SINDy algorithm for sparse learning of PDEs is based on

$$\arg\min_{\boldsymbol{\alpha}} \|\mathbf{V} - \mathbf{F}\boldsymbol{\alpha}\|_2^2 + \lambda \|\boldsymbol{\alpha}\|_1, \qquad (2.1)$$

in which $\mathbf{F}$ and $\mathbf{V}$ are the feature matrix and velocity vector, respectively. The vector $\boldsymbol{\alpha}$ contains the unknown coefficients appearing in the PDE and $\lambda$ is sparsity regularization parameter. An example of feature matrix $\mathbf{F}$, consisting of the field variable $u$ and its spatial derivatives in one dimensional space, $x$ can be constructed as

$$\mathbf{F} = \begin{bmatrix} 1 & u(x_1,t_1) & u_x(x_1,t_1) & \cdots & u_{xxx}^3(x_1,t_1) \\ 1 & u(x_2,t_1) & u_x(x_2,t_1) & \cdots & u_{xxx}^3(x_2,t_1) \\ 1 & u(x_3,t_1) & u_x(x_3,t_1) & \cdots & u_{xxx}^3(x_3,t_1) \\ \vdots & \vdots & \vdots & \ddots & \vdots \\ 1 & u(x_{n-1},t_m) & u_x(x_{n-1},t_m) & \cdots & u_{xxx}^3(x_{n-1},t_m) \\ 1 & u(x_n,t_m) & u_x(x_n,t_m) & \cdots & u_{xxx}^3(x_n,t_m) \end{bmatrix}, \qquad (2.2)$$

where each column represents a different feature at $n$ spatial points and $m$ time instances (snapshots); the first column consists of unit values to accommodate for the bias term of the solution, i.e., a potential constant source. The velocity vector $\mathbf{V}$ consisting of the time derivative of the field variable $u$ is expressed as

$$\mathbf{V} = \begin{bmatrix} u_t(x_1,t_1) \\ u_t(x_2,t_1) \\ u_t(x_3,t_1) \\ \vdots \\ u_t(x_{n-1},t_m) \\ u_t(x_n,t_m) \end{bmatrix}. \qquad (2.3)$$

The SINDy algorithm restates Eq. (2.1) as

$$\begin{aligned} \boldsymbol{\alpha}^0 &= \left(\mathbf{F}^T\mathbf{F}\right)^{-1}\mathbf{F}^T\mathbf{V} \\ S^k &= \left\{j \in [n_f] : |\boldsymbol{\alpha}_j^k| \geq \lambda\right\}, \quad k \geq 0 \\ \boldsymbol{\alpha}^{k+1} &= \arg\min_{\boldsymbol{\alpha} \in \mathbb{R}^{n_f} : \sup(\boldsymbol{\alpha}) \subseteq S^k} \|\mathbf{V} - \mathbf{F}\boldsymbol{\alpha}\|_2, \quad k \geq 0 \end{aligned} \qquad (2.4)$$

where $n_f$ is the number of features in the feature matrix. As shown by Zhang and Schaeffer [17], this algorithm presents attractive convergence features. However, it performs poorly in the presence of high correlation between the columns of feature library. On the other hand, STRidge algorithm is robust to the correlation between features. Therefore, Eq. (2.1) is modified to contain a penalty term, $\ell_2$ and it can be recast as

$$\arg\min_{\boldsymbol{\alpha}} \|\mathbf{V} - \mathbf{F}\boldsymbol{\alpha}\|_2^2 + \lambda_1 \|\boldsymbol{\alpha}\|_1 + \lambda_2 \|\boldsymbol{\alpha}\|_2, \qquad (2.5)$$

where $\lambda_1$ is the penalty parameter enforcing sparsity and $\lambda_2$ is the penalty parameter regularizing the magnitude of the recovered coefficients. The optimization of Eq. (2.5) can be achieved through the sequentially thresholded STRidge algorithm as

$$\begin{aligned}
\boldsymbol{\alpha}^0 &= \left(\mathbf{F}^T\mathbf{F} + \lambda_2 \mathbf{I}\right)^{-1} \mathbf{F}^T \mathbf{V} \\
S^k &= \left\{ j \in [n_f] : |\boldsymbol{\alpha}_j^k| \geq \lambda_1 \right\}, \quad k \geq 0 \\
\boldsymbol{\alpha}^{k+1} &= \arg\min_{\boldsymbol{\alpha} \in \mathbb{R}^{n_f} : \sup(\boldsymbol{\alpha}) \subseteq S^k} \|\mathbf{V} - \mathbf{F}\boldsymbol{\alpha}\|_2 + \lambda_2 \|\boldsymbol{\alpha}\|_2, \quad k \geq 0
\end{aligned} \qquad (2.6)$$

This algorithm has been used successfully to recover governing nonlinear partial differential equations [28]. However, the success of this algorithm diminishes significantly in presence of noise [28].

Fasel et al. [25] combined SINDy with ensemble learning to handle noisy measurements. Ensemble algorithms use a base learning algorithm and create multiple hypotheses. The resulting decision is obtained via aggregation of the outputs of the proposed hypotheses. These algorithms provide statistical, computational and representational advantages compared to the vanilla learning algorithms. Generally, resulting ensemble algorithm is significantly more accurate than the original classifier. However, to leverage the full potential of ensemble methods, unstable learning algorithms i.e., learning algorithms which are sensitive to small changes in training dataset, should be used.

Bagging (Bootstrap aggregating) [29] is one of the well known ensemble methods which is based on running the same learning method over different subsets of the same dataset. In Bagging, given a set of $M$ data points, $N$ distinct datasets are created by uniformly sampling $M$ data points with replacement. The original algorithm is trained on generated datasets and finally, hypotheses are combined with aggregating the trained models.

Fasel et al. [25] use bagging and bragging with Eq. (2.5) as the base learner. They demonstrate the capability of the proposed method by recovering the coefficients of several PDEs from noisy and scarce measurements. The instability of the learner is avoided by systematically removing features from the feature matrix and adding random noise to the dataset. The main algorithm for the ensemble SINDy approach is shown in Fig.1.

In this study, Ensemble SINDy approach is applied to learning 2D Fisher-Stefan system. This system involves a reaction-diffusion equation along with a moving boundary equation referred to as the Stefan condition. First equation evolves in a Cartesian coordinate system while the Stefan condition is defined in reference to the coordinate system of the moving boundary. Therefore, feature matrices for the discovery of Stefan condition is constructed using derivatives with respect to the moving boundary corotational coordinate system.

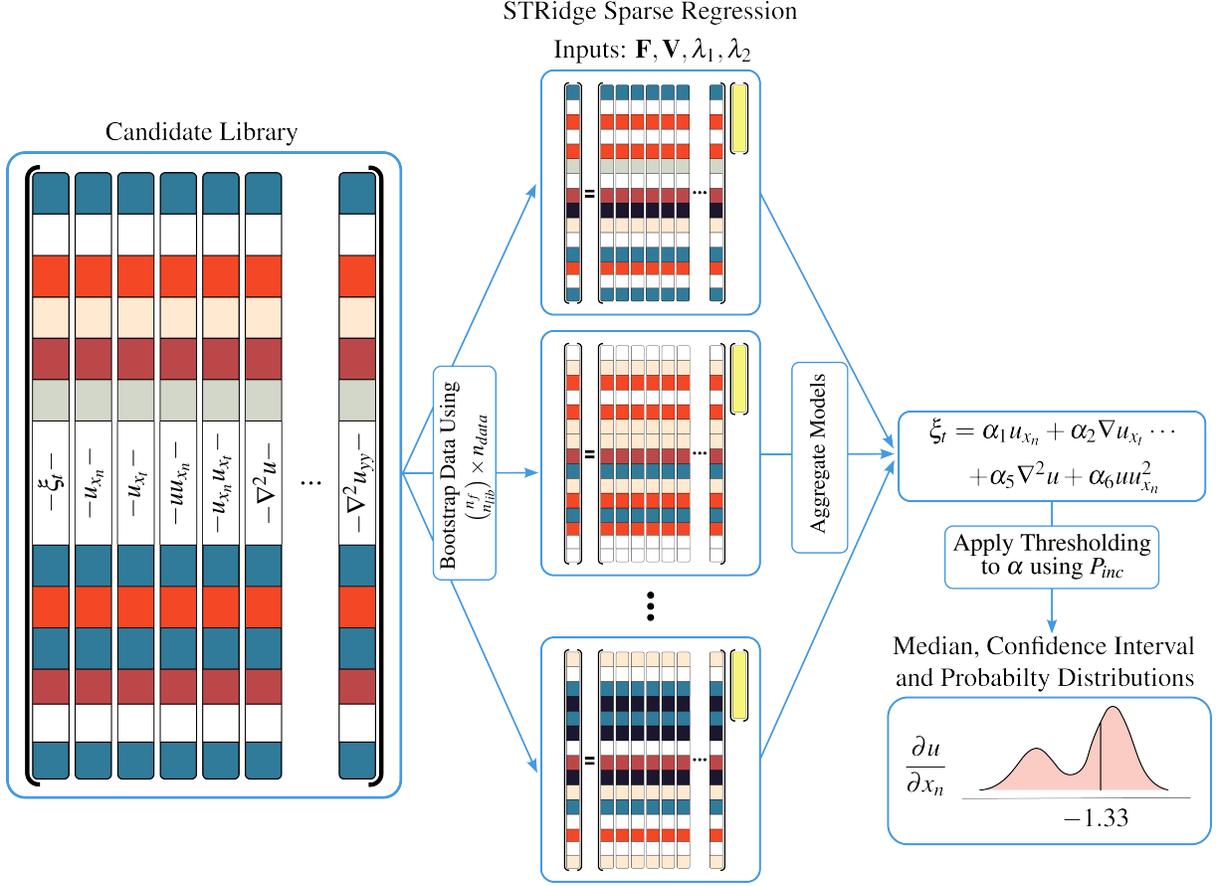

**Figure 1**. Ensemble SINDy algorithm

### 3. 2D Fisher-Stefan Problem

This study concerns the recovery of the governing equation of the 2D Fisher-Stefan model [30]. Tam et al. [31,32] recently conducted an extensive study on the 2D Fisher-Stefan model. This model has practical applications, such as representing cell tumor growth, fibroblast cells invading a partial wound or porous media modeling for population dynamics [33] making it tangible and important. We created the dataset using the open-source Julia code shared by Tam et al. [31,32]. 2D Fisher-Kolmogorov-Petrovsky-Piskunov (Fisher-KPP) equation is expressed as

$$\frac{\partial u}{\partial t} = \nabla \cdot (\nabla u) + u(1-u), \text{ for } \mathbf{x} \in \Omega_1 \qquad (3.1)$$

where $u(\mathbf{x},t)$ denotes the population density. This equation is a reaction-diffusion equation and has a logistic type of nonlinear source term on the right hand side $u(1-u)$ which controls the growth and competition of the species $u$ [34]. Inspired by the classical Stefan problem [8,35], Fisher-Stefan problem is defined by coupling Fisher-KPP equation with a moving boundary $\xi(t)$. This boundary moves in its normal direction defined by the gradient of the population density. As a result, the problem domain $\Omega_1$ evolves with the moving boundary. The Stefan and Dirichlet boundary conditions defined on the moving boundary are

$$\frac{\partial \xi_n}{\partial t} = -\kappa \nabla u \cdot \mathbf{n} \quad \text{and} \quad u = u_f \quad \text{for} \quad \mathbf{x} \in \xi(t) \tag{3.2}$$

where $\kappa > 0$ is the Stefan parameter, which connects the gradient of population density and surface evolution velocity. We consider two distinct test cases (datasets) for learning the governing equations of this system: a) a vertical moving boundary with a sinusoidal perturbance, and 2) a circular moving boundary with irregularity.

### 3.1 Vertical Moving Boundary with a Sinusoidal Perturbation
Spatial domain for this problem is defined as $\mathbf{x} \in [0,10] \times [0,10]$ and boundary conditions in $y$ direction are defined as periodic. Initial-boundary conditions for this problem are depicted in Fig. 2.

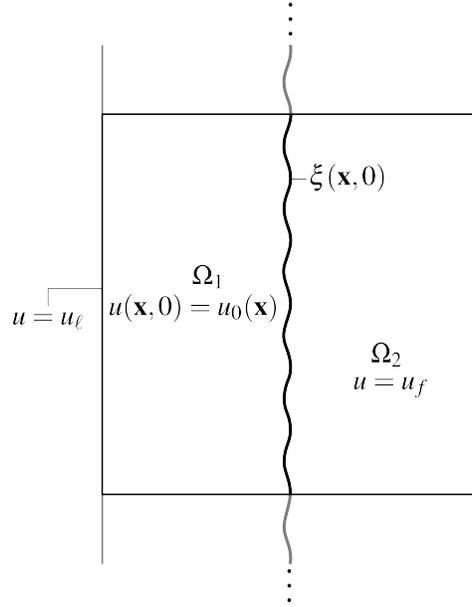

**Figure 2**. Initial-boundary conditions for the 2D Fisher-Stefan problem with a vertical moving boundary having a sinusoidal perturbation

The remaining boundary condition on the left edge representing a fixed population density and the periodic boundary conditions in $y$ direction are expressed as

$$\begin{aligned} u(0, y, t) &= u_\ell \\ u(x, y + nL_y, t) &= u(x, y, t), \quad \text{for} \quad n \in \mathbb{Z} \end{aligned} \tag{3.3}$$

The initial conditions are defined as

$$u(\mathbf{x},0) = u_0(\mathbf{x}), \text{ for } \mathbf{x} \in \Omega_1$$
$$u(\mathbf{x},0) = u_f, \text{ for } \mathbf{x} \in \Omega_2 \tag{3.4}$$

The solution to the system of equations is obtained using the open source code shared by [32] for $\kappa = 0.5$, $\Delta x = \Delta y = 0.1$, $\Delta t = 0.01$, $t \in [0,5]$ and the rest of the parameters of the problem can be also found in [32]. As time progresses, this solution for the population density along with the moving boundary is depicted in Fig. 3.

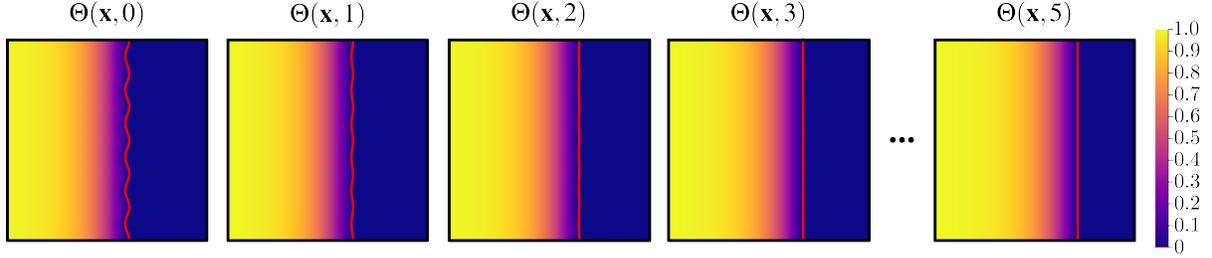

**Figure 3**. Time evolution of the field data $\Theta = [u, \xi]$ for the 2D Fisher-Stefan problem with a vertical moving boundary having a sinusoidal perturbation

### 3.2 Circular Moving Boundary with Irregularity

Figure 4 shows the spatial domain with $\mathbf{x} \in [-10,10] \times [-10,10]$ and the initial-boundary conditions,

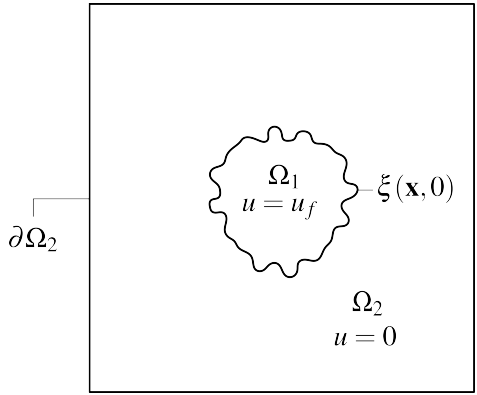

**Figure 4**. Initial-boundary conditions for the 2D Fisher-Stefan problem with a circular moving boundary having irregularity.

The initial position of the moving boundary $\xi(\mathbf{x},0)$ is defined as

$$\xi(\mathbf{x},0) = 3.5 + 0.2(\sin 3\theta + \sin 12\theta + \sin 16\theta), \tag{3.5}$$

where

$$\theta = \tan^{-1}\frac{y}{x}. \qquad (3.6)$$

The remaining boundary conditions on the outer edges representing fixed population density are expressed as

$$u(\mathbf{x},0) = 0 \text{ for } \mathbf{x} \in \partial\Omega_2 \qquad (3.7)$$

The initial conditions are defined as

$$\begin{aligned} u(\mathbf{x},0) &= u_f, \text{ for } \mathbf{x} \in \Omega_1 \\ u(\mathbf{x},0) &= 0, \text{ for } \mathbf{x} \in \Omega_2 \end{aligned} \qquad (3.8)$$

The solution to the system of equations is obtained using the open source code shared by Tam and Simpson [31] for $\kappa = 0.1$, $\Delta x = \Delta y = 0.1$, $\Delta t = 0.04$, $t \in [0,40]$. As time progresses, this solution for the population density along with the moving boundary is depicted in Fig. 5.

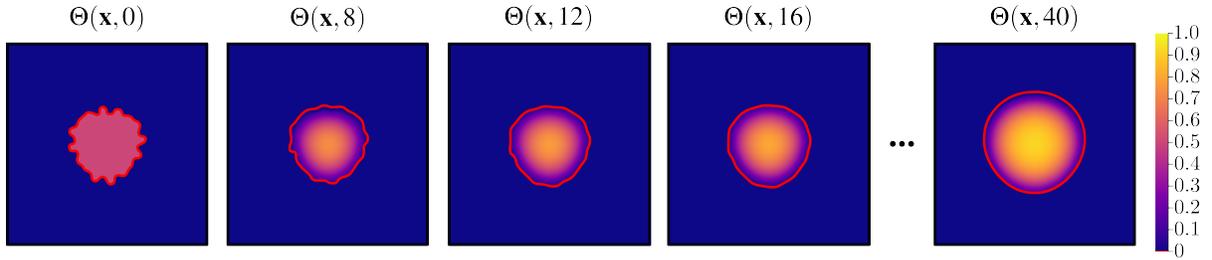

**Figure 5**. Time evolution of the field data $\Theta = [u,\xi]$ for the 2D Fisher-Stefan problem with a circular moving boundary having irregularity

### 3.3 Corotational Coordinate System of the Moving Boundary and the Surface Evolution Velocity

This section describes the learning of the underlying governing equations for the 2D Fisher-Stefan problem. The learning includes the 2D Fisher-KPP equation which evolves in a 2D Cartesian coordinate system and the governing equation for the moving boundary (Stefan condition) which evolves in its own corotational coordinate system.

The derivatives of the field are constructed by employing the PDDO [26, 41]. For the derivatives in Cartesian coordinate system, the approach described by Bekar and Madenci [27] is adopted without any special treatment (see Appendix for more details). The derivatives in corotational coordinates are calculated by rotating the coordinate system to the normal and tangent directions to the interface at the sensor locations. The families of the material points are also demarcated using the tangents. This approach is described in Fig. 6.

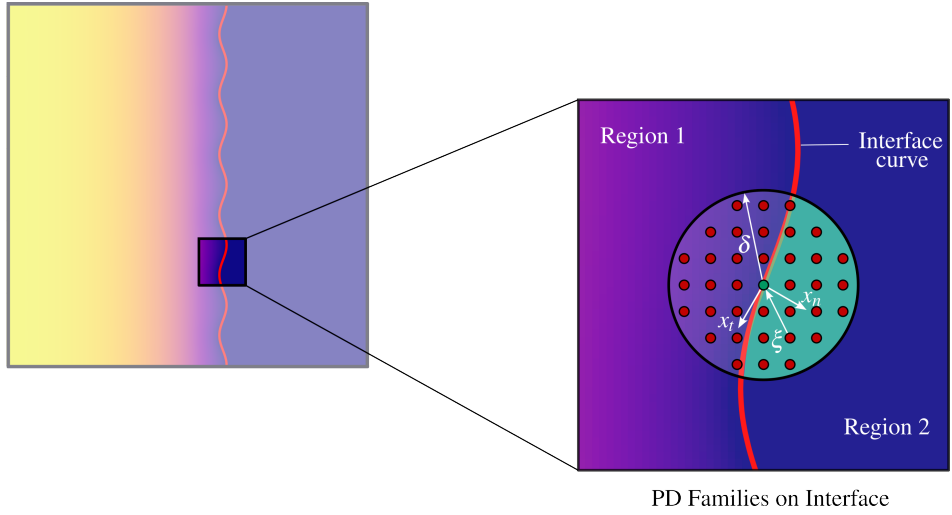

**Figure 6**. PD families and interface coordinate system on the moving boundary

Coordinate transformation can be accomplished by using two different methods. The first method is based on the use of arctangents of the line sections between sensor locations on the moving boundary. The second method is based on the use of level sets. Using the snapshots of the moving boundary, the level set equation is solved and the gradients of level sets are calculated. These gradients after normalization define the normal directions to the moving boundary (interface).

Fig. 7 illustrates the coordinate system rotation and advancing interface velocity calculations using panels between the coordinate locations on the interface. First, we consider a coordinate point, $\tilde{p}_i^t$ between two neighboring points on the moving interface. Subsequently, we rotate the coordinate system based on the arctangent of the angle $\theta$. At last, we calculate the distance vector $\mathbf{s}_i^t$ between the panel midpoint $\tilde{p}_i^t$ and the nearest coordinate point on the moving interface in the next time step. The norm of this distance vector can be divided to timestep size $\Delta t$ to calculate the velocity of the interface in the normal direction as

$$\left[\frac{\partial \xi_n}{\partial t}\right]^i \approx \frac{\left|\mathbf{s}_i^t\right|}{\Delta t} \qquad (3.9)$$

Alternative approach is to rotate the vector $\mathbf{s}_i^t$ using the calculated panel normal and use the norm of the resulting vector as

$$\left[\frac{\partial \xi_n}{\partial t}\right]^i \approx \frac{\left|\mathbf{s}_i^t \cdot \mathbf{n}_i^t\right|}{\Delta t} \qquad (3.10)$$

where $\mathbf{n}_i^t$ is the calculated panel normal at time $t$ and location $i$.

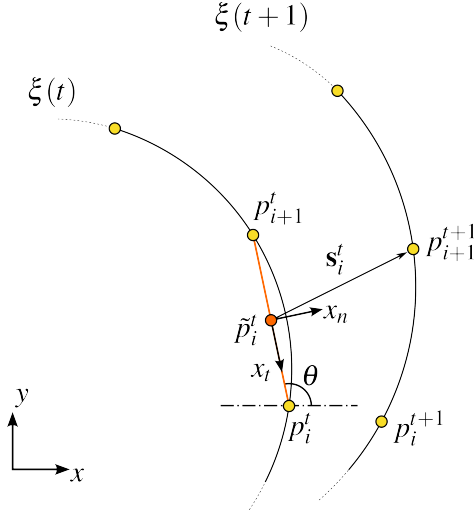

**Figure 7**. Discrete points on the moving boundary (interface), rotation of the coordinate system using panel angles and calculation of the distance vector

## 4. Numerical Results

This section presents the numerical results of the recovery experiments conducted to test the proposed learning framework. It presents the recovered models for the Stefan condition and Fisher-KPP equation, obtained for different levels of noise. The relative error of the model coefficients is calculated as

$$\varepsilon_c = \frac{\|\boldsymbol{\alpha} - \hat{\boldsymbol{\alpha}}\|_2}{\|\boldsymbol{\alpha}\|_2} \tag{4.1}$$

where $\boldsymbol{\alpha}$ is the vector of ground truth coefficients and $\hat{\boldsymbol{\alpha}}$ is the vector of the recovered coefficients. Our framework is implemented in Python, the codes and data are available at: https://github.com/alicanbekar/MB_PDDO-SINDy.

### 4.1 Discovery of the Stefan Condition

We employ Ensemble SINDy to discover the Stefan condition from the field measurements. The candidate features consisting of the field derivatives and their products are assembled as

$$\begin{bmatrix} 1 & u & u_{x_n} & u_{x_t} & u_{x_n} \times u_{x_n} & u_{x_n} \times u_{x_t} & u_{x_t} \times u_{x_t} & u_{x_n x_n} & u_{x_t x_t} & u_{x_n x_t} & u_{x_n x_n} + u_{x_t x_t} \end{bmatrix}. \tag{4.2}$$

The derivatives of the field variable $u$ are constructed in corotational coordinate system using the PDDO as described in detail in Appendix. This candidate space can also include derivatives of the field variable in the Cartesian coordinate system or higher order derivatives and their products. However, we keep the candidate space simple and assume that it includes only the terms to explain the underlying dynamics of the moving boundary. Furthermore, we normalize each feature by its maximum absolute value to prevent the optimizer to be biased.

We test Ensemble SINDy with library bagging on 3 different noise levels. We aggregate the models by choosing the median of the recovered coefficients from different bootstrapped datasets. Additionally, we construct the probability distributions using Gaussian kernel estimation [36]. For the dataset with no measurement noise, we bootstrap the data for 60 times and bootstrap the features by leaving 3 out at every regression. This results in 9900 different tests. We calculate the inclusion probabilities of the features by dividing the number of appearances to the total number of tests. We set the threshold for the inclusion probability $P_{inc}$ as 0.7. This means that we disregard the features appearing less than 70% of the time. We calculate the standard deviation of the coefficients of the features that appear more than $P_{inc}$, and construct the confidence intervals using the margins of $3\sigma$ from the mean $\mu$.

We chose the parameters for the STRidge regression as $\lambda_1 = 0.3$ and $\lambda_2 = 1.0$. The value of $\lambda_1$ can be chosen using information criteria like BIC [37] or AIC [38]. The Python library we share contains a function to evaluate the AIC scores of the models and returns the suitable value for $\lambda_1$.

### 4.1.1 Discovery of the Vertical Moving Boundary with a Sinusoidal Perturbation
Fig. 8 depicts the recovered coefficients, uncertainties, inclusion probabilities and the probability density estimations for the case with clean data.

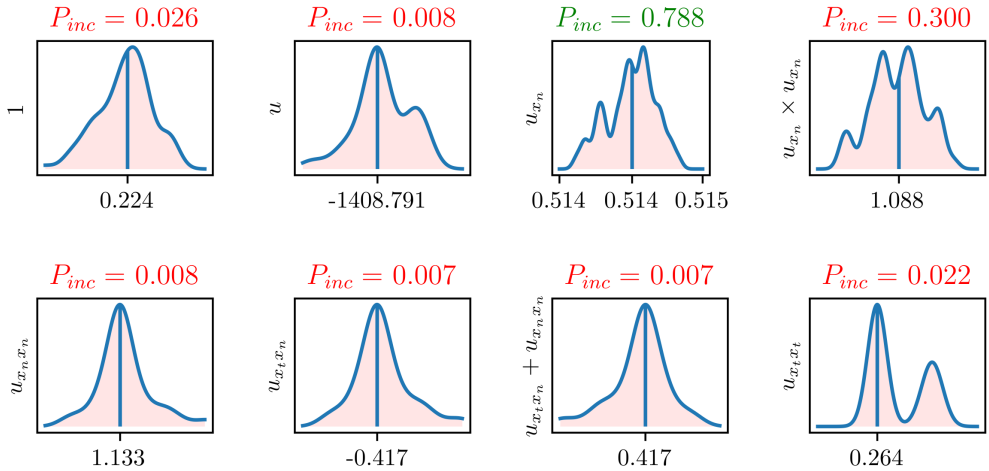

**Figure 8**. Median values, 99% $(\mu \pm 3\sigma)$ confidence intervals and inclusion probabilities for the Stefan condition with a vertical moving boundary having a sinusoidal perturbation for the clean dataset

Fig. 8 shows that out of 11 terms, only 8 terms appear at least once. Also, majority of the recovered coefficients appear less than 3% of the time. Only 2 terms have significant inclusion probabilities, $u_{x_n}$ and $u_{x_n} \times u_{x_n}$. Considering the correlation among them makes this appearance rational. The only term appearing more than the threshold probability is $u_{x_n}$ with $P_{inc} = 0.788$ and median value of 0.514, which is the term responsible for the underlying dynamics of the moving boundary. We calculate $\varepsilon_c$ using Eq. (4.1) as 0.028 which corresponds to less than 3% relative error in the recovered coefficient.

Fig. 8 also shows the 99% or $3\sigma$ confidence interval of the recovered coefficient of the feature $u_{x_n}$. We use the limits of the confidence interval and solve for the moving boundary location at $t = 2$ to show the uncertainty of the position of the interface. Fig. 9 depicts the region of uncertainty, ground truth solution and solution obtained with the median value from the recovery. Since the dataset is clean, the recovered coefficients and the position of the moving boundary has a narrow region of uncertainty. Furthermore, the solution obtained with the median value of the recovery coincides with the ground truth solution.

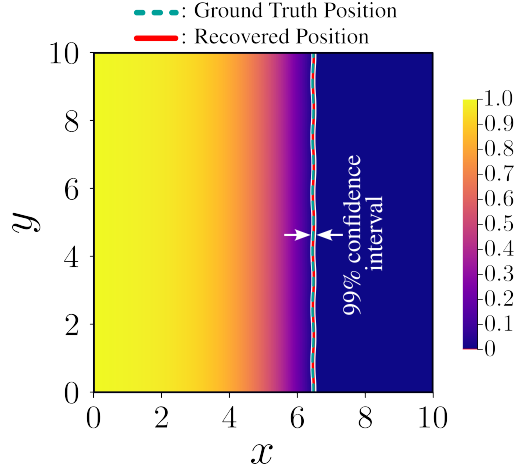

**Figure 9**. Interface positions with the confidence interval at $t = 2$ for the clean data case with a vertical moving boundary having a sinusoidal perturbation

Subsequently, we add noise to the training as $\bar{u} = u + n$ where $n \in \mathcal{N}(0, \sigma^2)$ and $\sigma$ is the standard deviation of $u$. The noise magnitude $\eta$ multiplied with $\sigma$ to introduce the desired level of noise to the dataset through the Gaussian randomness. The data is corrupted by considering 1% Gaussian random noise for the first trial. The learning and bragging parameters are the same as those of the clean data case for calculating the inclusion probabilities, confidence intervals and median values for the recovered coefficients. Fig. 10 shows the results for the recovery.

Fig. 10 shows that out of 11 terms, only 1 term appears more than 70% of the time. However, adding noise increases the probability of irrelevant features appearing in the recovered model. Also, the confidence interval of the coefficient of the feature $u_{x_n}$ increases and $3\sigma$ confidence interval corresponds to $[0.316, 0.617]$. Interestingly, inclusion probability of the correct model also increases for the case of 1% Gaussian noise with $P_{inc} = 0.975$ and median value of 0.504 is closer to the ground truth value compared to the case with clean data. This is caused by the probabilistic nature of the bagging process. However, noise in the field data increases the uncertainty of the recovered model, making the recovered median value less reliable. We calculate $\varepsilon_c$ as 0.008 which corresponds to less than 1% relative error in recovered coefficient. Fig. 11 depicts the region of uncertainty, ground truth solution and solution obtained with the median value from the recovery. The noise in the dataset widens the 99% confidence interval and increases the uncertainty of the predicted position of the moving boundary.

However, the recovered position using the recovered coefficient still overlaps with the ground truth position at $t = 2$.

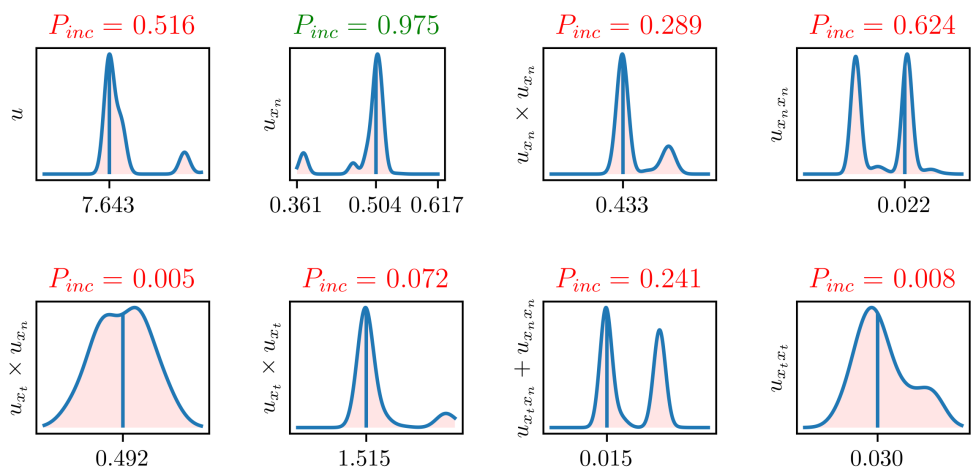

**Figure 10**. Median values, 99% $(\mu \pm 3\sigma)$ confidence intervals and inclusion probabilities for the Stefan condition with a vertical moving boundary having a sinusoidal perturbation using dataset with 1% noise level

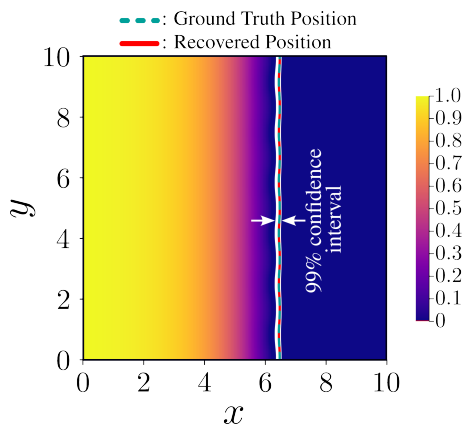

**Figure 11**. Interface position with the confidence interval at $t = 2$ for the case of 1% Gaussian noise case with a vertical moving boundary having a sinusoidal perturbation

Finally, we corrupt the data by considering 5% Gaussian random noise. The learning and bragging parameters are the same as those of the clean data case for calculating the inclusion probabilities, confidence intervals and median values for the recovered coefficients. Fig. 12 shows the results for the recovery. Out of 11 terms, 10 terms appear at least once. The $3\sigma$ confidence interval of the coefficient of the feature $u_{x_n}$ corresponds to $[-0.054, 0.732]$ with the median value of 0.387. The increased noise causes the recovered coefficient diverge from the ground truth value and irrelevant terms appear even more frequently. We calculate $\varepsilon_c$ as 0.226 which corresponds to ~23% relative error in recovered coefficient. However, Ensemble SINDy can properly identify the responsible terms for underlying dynamics even with the relatively high measurement noise.

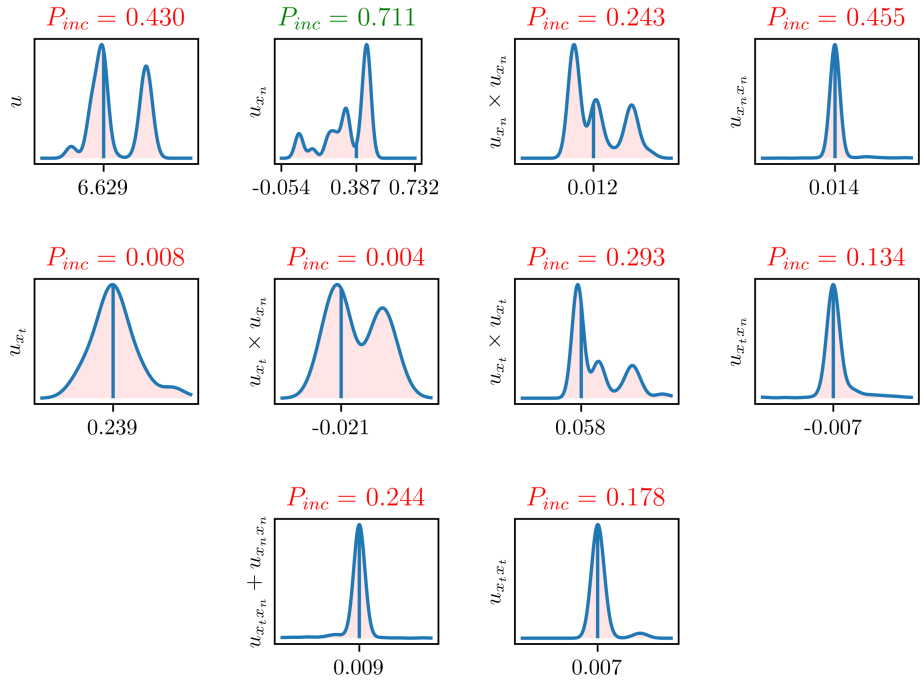

**Figure 12**. Median values, 99% $(\mu \pm 3\sigma)$ confidence intervals and inclusion probabilities for the Stefan condition with a vertical moving boundary having a sinusoidal perturbation using dataset with 5% noise level

Fig. 13 depicts the region of uncertainty, ground truth solution and solution obtained with the median value from the recovery. The noise in the dataset widens the 99% confidence interval even more and increases the uncertainty of the predicted position of the moving boundary, making the recovered position less reliable. Furthermore, the discrepancy between the ground truth position and the recovered position increases. However, compared to the confidence region, recovered position is relatively close to the ground truth position. Table 1 shows the comparison of the recovered coefficients with the ground truth model.

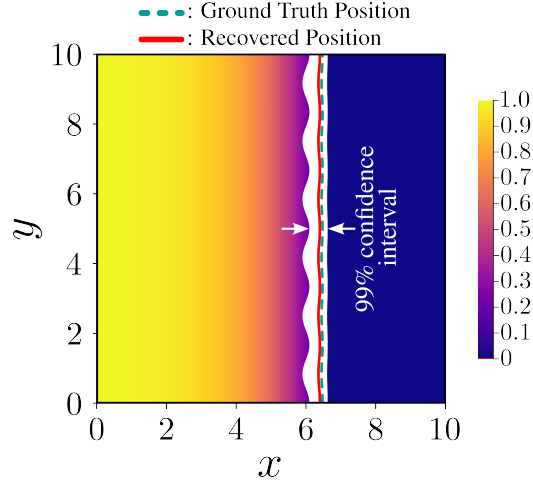

**Figure 13**. Interface positions with the confidence interval at $t=2$ for the 5% Gaussian noise case with a vertical moving boundary having a sinusoidal perturbation

**Table 1**. Identified models and corresponding confidence intervals for Stefan condition with a vertical moving boundary having a sinusoidal perturbation

|  | Model | Confidence Interval |
|---|---|---|
| Ground truth | $\dfrac{\partial \xi_n}{\partial t} = -0.5 u_{x_n}$ | – |
| Clean data | $\dfrac{\partial \xi_n}{\partial t} = -0.514 u_{x_n}$ | $[0.514, 0.515]$ |
| Identified model (1% noise) | $\dfrac{\partial \xi_n}{\partial t} = -0.504 u_{x_n}$ | $[0.316, 0.617]$ |
| Identified model (5% noise) | $\dfrac{\partial \xi_n}{\partial t} = -0.387 u_{x_n}$ | $[-0.054, 0.732]$ |

**4.1.2 Discovery for Circular Moving Boundary with Irregularity**

For this case, we modify the sparsity penalty and change it from $\lambda_1 = 0.3$ to $\lambda_1 = 0.1$. The reason for this modification is that none of the inclusion probabilities exceed the threshold inclusion probability for $\lambda_1 = 0.3$. Fig. 14 depicts the recovered coefficients, uncertainties, inclusion probabilities and the probability density estimations for the case with the clean dataset.

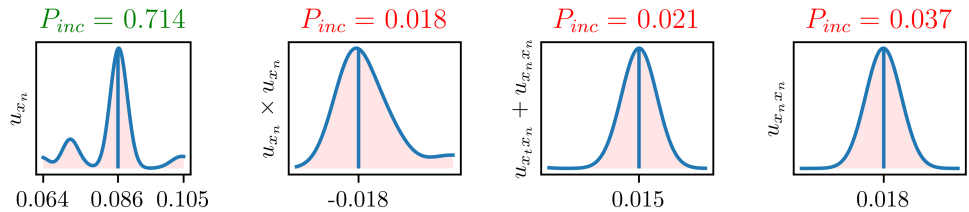

**Figure 14**. Median values, 99% $(\mu \pm 3\sigma)$ confidence intervals and inclusion probabilities for the Stefan condition with a circular moving boundary having irregularity using the clean dataset

Fig. 14 shows that out of 11 terms, only 4 terms appear at least once. Also, majority of the recovered coefficients appear less than 4% of the time. Only 1 term has significant inclusion probability, $u_{x_n}$, appearing more than the threshold probability with $P_{inc} = 0.714$ and median value of 0.086. Thus, it is the term responsible for the underlying dynamics of the moving boundary. We calculate $\varepsilon_c$ as 0.14 which corresponds to less than 15% relative error in the recovered coefficient.

Fig. 14 also shows the 99% or $3\sigma$ confidence interval of the recovered coefficient of the feature $u_{x_n}$. We use the limits of the confidence interval and solve for the moving boundary location at $t = 40$ to show the uncertainty of the position of the interface. Fig. 15 depicts the region of uncertainty, ground truth solution and solution obtained with the median value from the recovery. Compared to Fig. 9, this solution has a broader region of uncertainty. This is primarily caused by the longer duration of the analysis.

Subsequently, we corrupt the data using 1% Gaussian random noise. Confidence intervals and median values for the recovered coefficients are depicted in Fig. 16. Fig. 16 shows that out of 11 terms, only 1 appears more than 70% of the time. However, adding noise increases the probability of irrelevant features appearing in the recovered model. Additionally, the confidence interval of the coefficient of the feature $u_{x_n}$ increases and $3\sigma$ confidence interval corresponds to $[0.037, 0.166]$. However, the median value of 0.105 is closer to the ground truth value compared to the case with the clean data. We calculate $\varepsilon_c$ as 0.05 which corresponds to ~5% relative error in the recovered coefficient.

Fig. 17 depicts the region of uncertainty, ground truth solution and solution obtained with the median value from the recovery. Similar to the previous case, the noise in the dataset widens the 99% confidence interval and increases the uncertainty of the predicted position of the moving boundary. However, recovered position using the median value of the recovered coefficient still overlaps with the ground truth position at $t = 40$.

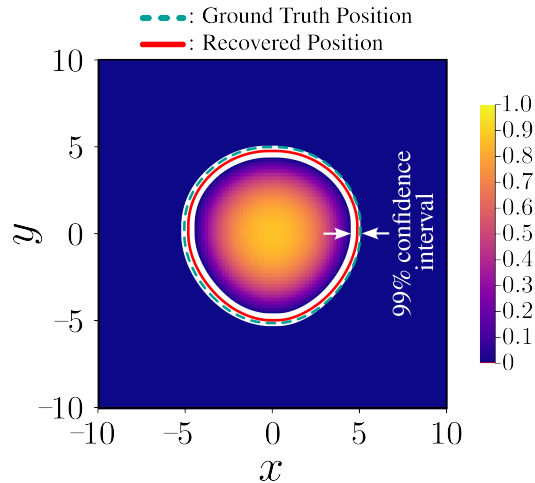

**Figure 15**. Interface positions with the confidence interval at $t = 2$ for the clean data case with a circular moving boundary having irregularity

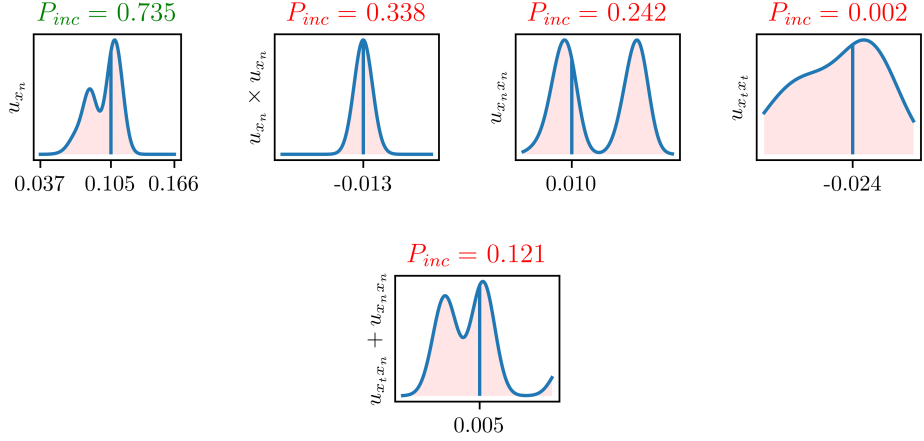

**Figure 16.** Median values, 99% $(\mu \pm 3\sigma)$ confidence intervals and inclusion probabilities for the Stefan condition with a circular moving boundary having irregularity using dataset with 1% noise level

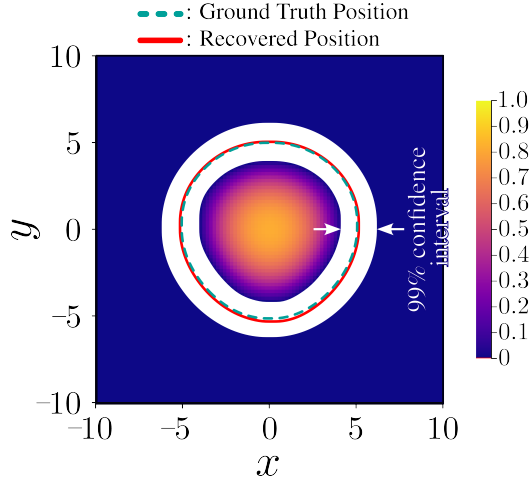

**Figure 17.** Interface positions with the confidence interval at $t = 2$ for the 1% Gaussian noise case with a circular moving boundary having irregularity

Finally, we corrupt the data using 5% Gaussian random noise. The learning and bragging parameters are the same as those of the clean data case for calculating the inclusion probabilities, confidence intervals and median values for the recovered coefficients. Fig. 18 shows the results for the recovery. Out of 11 terms, 10 terms appear at least once, 2 terms appear with high inclusion probabilities. Similar to the clean data case of the vertical moving boundary having a sinusoidal perturbation, the term $u_{x_n} \times u_{x_n}$ appears in addition to the term $u_{x_n}$. The $3\sigma$ confidence interval of the coefficient of the feature $u_{x_n}$ corresponds to $[0.048, 0.204]$ with the median value of 0.106. We calculate $\varepsilon_c$ as 0.06 which corresponds to ~6% relative error in recovered coefficient.

Fig. 19 depicts the region of uncertainty, ground truth solution and solution obtained with the median value from the recovery. The noise in the dataset widens the 99% confidence interval even more and increases the uncertainty of the predicted position of the moving boundary, making the recovered position less reliable. Table 2 shows the comparison of the recovered coefficients with the ground truth model.

**Table 2**. Identified models and corresponding confidence intervals for Stefan condition with a circular moving boundary having irregularity

|  | Model | Confidence Interval |
|---|---|---|
| Ground truth | $\frac{\partial \xi_n}{\partial t} = -0.1 u_{x_n}$ | – |
| Clean data | $\frac{\partial \xi_n}{\partial t} = -0.086 u_{x_n}$ | $[0.064, 0.105]$ |
| Identified model (1% noise) | $\frac{\partial \xi_n}{\partial t} = -0.105 u_{x_n}$ | $[0.037, 0.166]$ |
| Identified model (5% noise) | $\frac{\partial \xi_n}{\partial t} = -0.106 u_{x_n}$ | $[0.048, 0.204]$ |

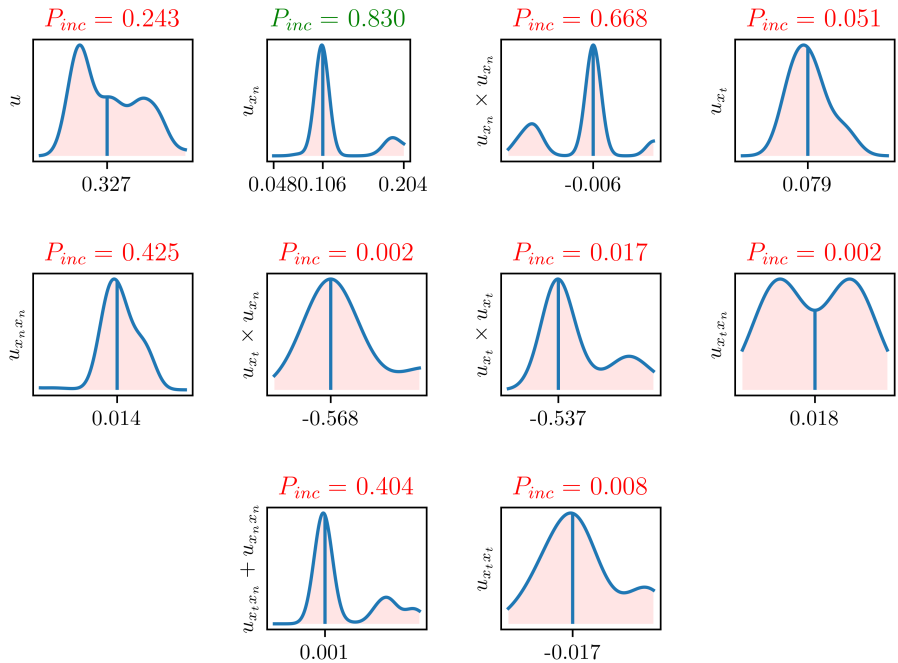

**Figure 18**. Median values, 99% $(\mu \pm 3\sigma)$ confidence intervals and inclusion probabilities for the Stefan condition with a circular moving boundary having irregularity using dataset with 5% noise level

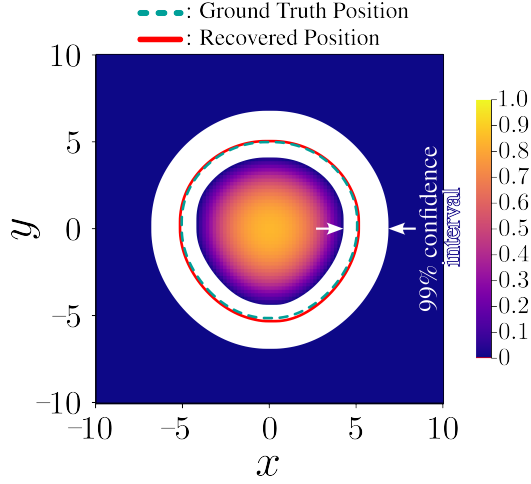

**Figure 19**. Interface positions with the confidence interval at $t = 2$ for the 5% Gaussian noise case with a circular moving boundary having irregularity

### 4.2 Discovery of the Fisher-KPP Equation

For this case, we also employ Ensemble SINDy to discover the Fisher-KPP equation from the field measurements. Our candidate space terms for the discovery are based on the assumption that the population density evolves in a Cartesian coordinate system. Therefore, the derivatives of the field variable $u$ are constructed in $x$ and $y$ directions. This is the inductive bias of our model for the Fisher-KPP equation. The candidate space consisting the field derivatives and their products is assembled as

$$\begin{bmatrix} 1 & u & u^2 & \nabla^2 u & u_x & u_y & u_{xx} & u_{xy} & u_{yy} & u_y u_{xx} \end{bmatrix} \quad (4.3)$$

Again, we keep the candidate space simple and assume that it includes only the terms to explain the underlying dynamics of the Fisher-KPP equation. We scale each feature by its maximum absolute value to prevent the optimizer to be biased.

Fisher-KPP equation with the chosen initial conditions is a challenging case for discovery. The main reason for this is that the saturation of the concentration values is far from the moving boundary. When the field variable saturates, its derivatives vanish, giving no information about the field equation; nevertheless, we use Ensemble SINDy with library bagging on Fisher-KPP equation to discover the underlying dynamics. For the dataset with no measurement noise, we bootstrap the data for 60 times and bootstrap the features by leaving 1 out at every regression. This results in 600 different tests. We set the threshold for the inclusion probability $P_{inc}$ as 0.7. We calculate the standard deviation of the coefficients of the features appearing more than $P_{inc}$ and construct the confidence interval using $3\sigma$. The learning and parameters are specified as the same as those of the Stefan condition experiments. Fig. 20 shows the results for the recovery for Fisher-KPP equation.

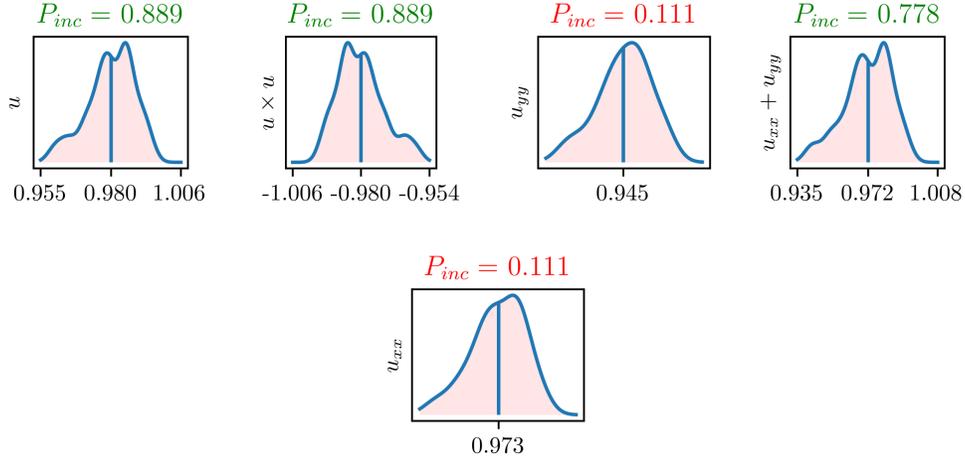

**Figure 20**. Median values, 99% $(\mu \pm 3\sigma)$ confidence intervals and inclusion probabilities for the Fisher-KPP equation using the clean dataset

Fig. 20 shows that out of 10 terms, only 5 terms appear at least once. The terms $u$ and $u^2$ appear with $P_{inc} = 0.889$ and median values have the same magnitude $\pm 0.980$ with opposite signs. This means that these terms appear as a pair and can be combined to a form $u(1-u)$. Laplacian term $\nabla^2 u$ appears with $P_{inc} = 0.778$ which is also greater than the threshold probability with the median value 0.972. Other appearing terms $u_{xx}$ and $u_{yy}$ also show up in pairs when Laplacian term is singled out, which is combined to form the Laplacian term. Hence, all recovered coefficients are relevant but the algorithm chooses the parsimonious solution. We calculate $\varepsilon_c$ as 0.022 which corresponds to less than 3% relative error in the recovered coefficients. Fig. 21 shows the absolute error between the ground truth solution and the recovered solution with the largest error being $\approx 4 \times 10^{-3}$.

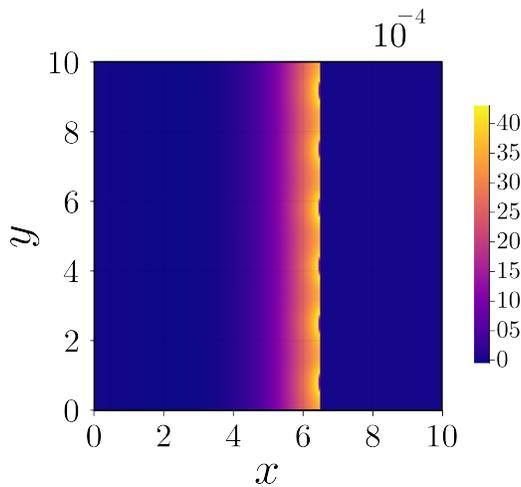

**Figure 21**. The absolute error between the ground truth and the recovered solutions

Additional Gaussian noise in Fisher-KPP equation results in recovering incorrect models, vanishing derivatives might be the real cause of this failure. Weak SINDy can be employed for this problem to increase recovery success. However, Weak SINDy is beyond the scope of this study.

## 5. Discussion and Conclusions

This study proposes a novel framework for learning underlying physics of processes with moving boundaries. By combining the recently introduced Ensemble SINDy and PDDO, we have successfully recovered the moving boundary equation and Fisher-KPP equation of Fisher-Stefan model. We impose the inductive bias assuming that the corotational coordinate system admits the parsimonious dynamical model. While this inductive bias does not limit the additional features evolving in a Cartesian coordinate system, it is a critical part of the algorithm. We have demonstrated the robustness of the present approach by considering various levels of noise in the measured data. We present the confidence intervals of recovered coefficients and demonstrate the uncertainties of the moving boundary positions by obtaining the solutions using the recovered coefficients.

Although the main focus is Fisher-Stefan model, this proposed framework is applicable to any kind of moving boundary problem with a smooth interface without a mushy region between the interacting domains. There are many other potential applications of the proposed framework, but one interesting application can be the discovery of mathematical models for regrowing limbs and tissue [39,40]. Regrowing limbs by nature is a moving boundary problem, and mathematical understanding this complex mechanism can help researchers govern or predict the process better.

In spite of the success of the proposed framework for learning physics with moving boundaries, there is still room for improvement. One limitation is locating the moving boundary. We assume that the location of the moving boundary is known beforehand. This process can be automated using a segmentation algorithm. Another extension of the proposed algorithm can be discovery of parametric PDEs. Ultimately, proposed algorithm might also be integrated with the Weak SINDy approach, this can allow the recovery of Fisher-KPP equation in presence of noise. A possible extension of the Weak SINDy can be by using the weak form of peridynamics for discovering the nonlocal forms of the local PDEs [41]. The assumption of the parsimonious dynamics evolve in moving boundary curve corotational coordinate system can also be relaxed using an approach similar to [18].

## Acknowledgments

EM and ACB performed this work as part of the ongoing research at the MURI Center for Material Failure Prediction through Peridynamics at the University of Arizona (AFOSR Grant No. FA9550-14-1-0073).

## Appendix - Peridynamic Differential Operator

In a 2-dimensional space, a function, $f(\mathbf{x}+\boldsymbol{\xi})$ can be expressed in terms of Taylor Series Expansion (TSE) as

$$f(\mathbf{x}+\boldsymbol{\xi}) = f(\mathbf{x}) + \xi_1 \frac{\partial f(\mathbf{x})}{\partial x_1} + \xi_2 \frac{\partial f(\mathbf{x})}{\partial x_2} + \frac{1}{2!}\xi_1^2 \frac{\partial^2 f(\mathbf{x})}{\partial x_1^2} + \frac{1}{2!}\xi_2^2 \frac{\partial^2 f(\mathbf{x})}{\partial x_2^2} + \xi_1\xi_2 \frac{\partial^2 f(\mathbf{x})}{\partial x_1 \partial x_2} + \mathcal{R} \quad \text{(A.1)}$$

where $\mathcal{R}$ is the remainder. Multiplying each term with PD functions, $g_2^{p_1 p_2}(\xi)$ and integrating over the domain of interaction (family), $\mathbf{H}_\mathbf{x}$ result in

$$\int_{\mathbf{H}_\mathbf{x}} f(\mathbf{x}+\xi)\, g_2^{p_1 p_2}(\xi)dV = \int_{\mathbf{H}_\mathbf{x}} f(\mathbf{x})\, g_2^{p_1 p_2}(\xi)dV + \frac{\partial f(\mathbf{x})}{\partial x_1}\int_{\mathbf{H}_\mathbf{x}} \xi_1 g_2^{p_1 p_2}(\xi)dV$$
$$+ \frac{\partial f(\mathbf{x})}{\partial x_2}\int_{\mathbf{H}_\mathbf{x}} \xi_2 g_2^{p_1 p_2}(\xi)dV + \frac{1}{2}\frac{\partial f^2(\mathbf{x})}{\partial x_1^2}\int_{\mathbf{H}_\mathbf{x}} \xi_1^2 g_2^{p_1 p_2}(\xi)dV \quad (A.2)$$
$$+ \frac{1}{2}\frac{\partial f^2(\mathbf{x})}{\partial x_2^2}\int_{\mathbf{H}_\mathbf{x}} \xi_2^2 g_2^{p_1 p_2}(\xi)dV + \frac{\partial^2 f(\mathbf{x})}{\partial x_1 \partial x_2}\int_{\mathbf{H}_\mathbf{x}} \xi_1 \xi_2 g_2^{p_1 p_2}(\xi)dV$$

in which the point $\mathbf{x}$ is not necessarily symmetrically located in the domain of interaction. The initial relative position, $\xi$, between points $\mathbf{x}$ and $\mathbf{x}'$ can be expressed as $\xi = \mathbf{x}' - \mathbf{x}$. This ability permits each point to have its own unique family with an arbitrary position. Therefore, the size and shape of each family can be different, and they significantly influence the degree of nonlocality. In general, the family of a point can be nonsymmetric due to nonuniform spatial discretization.

The degree of interaction between the material points in each family is specified by a nondimensional weight function, $w(|\xi|)$, which can vary from point to point. The weight function is usually chosen as Gaussian as $w(|\xi|) = e^{-4|\xi|^2/\delta^2}$ and this weight function and the interaction domain can be shown as

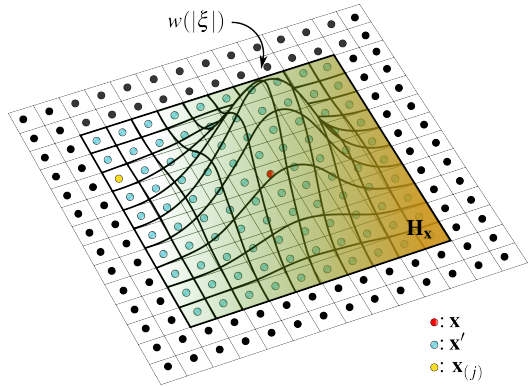

**Figure 22**. Interaction domain $\mathbf{H}_\mathbf{x}$ of point $\mathbf{x}$ and the form of the Gaussian weight function

The interactions become more local with decreasing family size. Thus, the family size and shape are important parameters. Each point occupies an infinitesimally small entity such as volume, area or a distance. The PD functions are constructed such that they are orthogonal to each term in the TSE as

$$\frac{1}{n_1! n_2!}\int_{\mathbf{H}_\mathbf{x}} \xi_1^{n_1} \xi_2^{n_2} g_2^{p_1 p_2}(\xi)dV = \delta_{n_1 p_1}\delta_{n_2 p_2} \quad (A.3)$$

with $(n_1, n_2, p, q = 0, 1, 2)$ and $\delta_{ij}$ is the Kronecker delta symbol. Enforcing the orthogonality conditions in the TSE leads to the nonlocal PD representation of the function itself and its derivatives as

$$f(\mathbf{x}) = \int_{H_x} f(\mathbf{x}+\xi) g_2^{00}(\xi) dV \tag{A.4a}$$

$$\begin{Bmatrix} \dfrac{\partial f(\mathbf{x})}{\partial x} \\ \dfrac{\partial f(\mathbf{x})}{\partial y} \end{Bmatrix} = \int_{H_x} f(\mathbf{x}+\xi) \begin{Bmatrix} g_2^{10}(\xi) \\ g_2^{01}(\xi) \end{Bmatrix} dV \tag{A.4b}$$

$$\begin{Bmatrix} \dfrac{\partial^2 f(\mathbf{x})}{\partial x^2} \\ \dfrac{\partial^2 f(\mathbf{x})}{\partial y^2} \\ \dfrac{\partial^2 f(\mathbf{x})}{\partial x \partial y} \end{Bmatrix} = \int_{H_x} f(\mathbf{x}+\xi) \begin{Bmatrix} g_2^{20}(\xi) \\ g_2^{02}(\xi) \\ g_2^{11}(\xi) \end{Bmatrix} dV \tag{A.4c}$$

The PD functions can be constructed as a linear combination of polynomial basis functions

$$g_2^{p_1 p_2} = a_{00}^{p_1 p_2} w_{00}(|\xi|) + a_{10}^{p_1 p_2} w_{10}(|\xi|) \xi_1 + a_{01}^{p_1 p_2} w_{01}(|\xi|) \xi_2 + a_{20}^{p_1 p_2} w_{20}(|\xi|) \xi_1^2 \\ + a_{02}^{p_1 p_2} w_{02}(|\xi|) \xi_2^2 + a_{11}^{p_1 p_2} w_{11}(|\xi|) \xi_1 \xi_2, \tag{A.5}$$

where $a_{q_1 q_2}^{p_1 p_2}$ are the unknown coefficients, $w_{q_1 q_2}(|\xi|)$ are the influence functions, and $\xi_1$ and $\xi_2$ are the components of the vector $\xi$. Assuming $w_{q_1 q_2}(|\xi|) = w(|\xi|)$ and incorporating the PD functions into the orthogonality equation lead to a system of algebraic equations for the determination of the coefficients as

$$\mathbf{A}\mathbf{a} = \mathbf{b}, \tag{A.6}$$

Where

$$\mathbf{A} = \int_{H_x} w(|\xi|) \begin{bmatrix} 1 & \xi_1 & \xi_2 & \xi_1^2 & \xi_2^2 & \xi_1 \xi_2 \\ \xi_1 & \xi_1^2 & \xi_1 \xi_2 & \xi_1^3 & \xi_1 \xi_2^2 & \xi_1^2 \xi_2 \\ \xi_2 & \xi_1 \xi_2 & \xi_2^2 & \xi_1^2 \xi_2 & \xi_2^3 & \xi_1 \xi_2^2 \\ \xi_1^2 & \xi_1^3 & \xi_1^2 \xi_2 & \xi_1^4 & \xi_1^2 \xi_2^2 & \xi_1^3 \xi_2 \\ \xi_2^2 & \xi_1 \xi_2^2 & \xi_2^3 & \xi_1^2 \xi_2^2 & \xi_2^4 & \xi_1 \xi_2^3 \\ \xi_1 \xi_2 & \xi_1^2 \xi_2 & \xi_1 \xi_2^2 & \xi_1^3 \xi_2 & \xi_1 \xi_2^3 & \xi_1^2 \xi_2^2 \end{bmatrix} dV, \tag{A.7a}$$

$$\mathbf{a} = \begin{bmatrix} a_{00}^{00} & a_{10}^{00} & a_{01}^{00} & a_{20}^{00} & a_{02}^{00} & a_{11}^{00} \\ a_{00}^{10} & a_{10}^{10} & a_{01}^{10} & a_{20}^{10} & a_{02}^{10} & a_{11}^{10} \\ a_{00}^{01} & a_{10}^{01} & a_{01}^{01} & a_{20}^{01} & a_{02}^{01} & a_{11}^{01} \\ a_{00}^{20} & a_{10}^{20} & a_{01}^{20} & a_{20}^{20} & a_{02}^{20} & a_{11}^{20} \\ a_{00}^{02} & a_{10}^{02} & a_{01}^{02} & a_{20}^{02} & a_{02}^{02} & a_{11}^{02} \\ a_{00}^{11} & a_{10}^{11} & a_{01}^{11} & a_{20}^{11} & a_{02}^{11} & a_{11}^{11} \end{bmatrix},$$ (A.7b)

And

$$\mathbf{b} = \begin{bmatrix} 1 & 0 & 0 & 0 & 0 & 0 \\ 0 & 1 & 0 & 0 & 0 & 0 \\ 0 & 0 & 1 & 0 & 0 & 0 \\ 0 & 0 & 0 & 2 & 0 & 0 \\ 0 & 0 & 0 & 0 & 2 & 0 \\ 0 & 0 & 0 & 0 & 0 & 1 \end{bmatrix}.$$ (A.7c)

After determining the coefficients $a_{q_1 q_2}^{p_1 p_2}$ via $\mathbf{a} = \mathbf{A}^{-1}\mathbf{b}$, the PD functions $g_2^{p_1 p_2}(\boldsymbol{\xi})$ can be constructed. The detailed derivations and the associated computer programs can be found in [41]. The PDDO is nonlocal; however, in the limit as the horizon size approaches zero, it recovers the local differentiation as proven by Silling and Lehoucq [42].